\documentclass[letterpaper, 10 pt, conference]{ieeeconf}  %

\IEEEoverridecommandlockouts                              %

\usepackage{xcolor}
\usepackage[bookmarks=true]{hyperref}
\usepackage[natbib=true,
    style=ieee,
    citestyle=numeric-comp,
    backend=biber,
    maxbibnames=10,
    maxcitenames=99,
    doi=false,
    url=false,
    giveninits=true]{biblatex}
\renewcommand*{\bibfont}{\footnotesize}

\usepackage{amssymb}
\usepackage{amsmath}
\usepackage{multirow}
\usepackage{multicol}
\usepackage{balance}
\usepackage{siunitx}

\usepackage[ruled,vlined,linesnumbered]{algorithm2e}
\SetInd{0.1em}{0.4em}
\SetAlFnt{\footnotesize\sf}
\SetKwRepeat{Do}{do}{while}
\SetKw{KwStep}{step}
\SetKwInput{KwData}{Input}
\SetKwInput{KwResult}{Result}
\SetKwBlock{RepeatForever}{repeat}{}
\SetKw{Continue}{continue}
\SetKwComment{Comment}{$\triangleright$\ }{}
\SetCommentSty{itshape}

\usepackage{tikz}
\usetikzlibrary{fit,calc}
\newcommand*{\tikzmk}[1]{\tikz[remember picture,overlay,] \node (#1) {};\ignorespaces}

\newcommand{\marklineTwo}[1]{\tikz[remember picture,overlay]{\node[yshift=2pt,xshift=#1,fill=yellow!100,opacity=.25,fit={(A)($(A)+(0.95\linewidth,-1.3\baselineskip)$)},rounded corners=4pt] {};}\ignorespaces}

\newcommand{\marklineOne}[1]{\tikz[remember picture,overlay]{\node[yshift=2pt,xshift=#1,fill=yellow!100,opacity=.25,fit={(A)($(A)+(0.95\linewidth,-0.3\baselineskip)$)},rounded corners=4pt] {};}\ignorespaces}

\usepackage[capitalise]{cleveref} %

\newcommand{\x}{\mathbf{x}}
\newcommand{\xdot}{\dot{\mathbf{x}}}
\newcommand{\X}{\mathcal{X}}
\newcommand{\Xseq}{\mathbf{X}}
\newcommand{\Useq}{\mathbf{U}}

\newcommand{\ui}{\mathbf{u}}
\newcommand{\U}{\mathcal{U}}

\newcommand{\W}{\mathcal{W}}
\newcommand{\Rdx}{\mathbb{R}^{d_{x}}}
\newcommand{\Rdxi}{\mathbb{R}^{d_{x}^{i}}}
\newcommand{\Rdu}{\mathbb{R}^{d_{u}}}
\newcommand{\Rdui}{\mathbb{R}^{d_{u}^{i}}}

\newcommand{\C}{\mathcal{C}}

\newcommand{\lcki}{l_{c_k}^i}

\newcommand{\px}{p_x}

\newcommand{\py}{p_y}

\newcommand{\thi}{\theta}

\newcommand{\A}{\mathbf{A}}

\newcommand{\p}{\mathbf{p}}
\newcommand{\R}{\mathbf{R}}
\newcommand{\wdot}{\dot{\boldsymbol{\Omega}}}
\newcommand{\wuav}{\boldsymbol{\Omega}}
\newcommand{\wuavfirst}{\boldsymbol{\Omega}^1}
\newcommand{\wuavn}{\boldsymbol{\Omega}^N}
\newcommand{\ez}{\mathbf{e}_3}

\newcommand{\po}{\mathbf{p}^0}
\newcommand{\poest}{\mathbf{p}^{0_e}}

\newcommand{\vo}{\dot{{\mathbf{p}}}^0}

\newcommand{\sci}{\mathbf{s}^i}
\newcommand{\scfirst}{\mathbf{s}^1}
\newcommand{\scn}{\mathbf{s}^N}

\newcommand{\scidot}{\dot{\mathbf{s}}^i}
\newcommand{\scihat}{\hat{\mathbf{s}}^i}

\newcommand{\wi}{\boldsymbol{\omega}^i}
\newcommand{\wifirst}{\boldsymbol{\omega}^1}
\newcommand{\win}{\boldsymbol{\omega}^N}

\newcommand{\podd}{\ddot{\mathbf{p}}^{0}}

\newcommand{\dbAstar}{\text{db-A}^\ast}

\title{\LARGE \bf
pc-dbCBS: Kinodynamic Motion Planning of Physically-Coupled Robot Teams
}

\author{Khaled Wahba and Wolfgang Hönig}

\begin{document}

\maketitle
\thispagestyle{empty}
\pagestyle{empty}

\begin{abstract}

Motion planning problems for physically-coupled multi-robot systems in cluttered environments are challenging due to their high dimensionality. 
Existing methods combining sampling-based planners with trajectory optimization produce suboptimal results and lack theoretical guarantees.

We propose Physically-coupled discontinuity-bounded Conflict-Based Search (pc-dbCBS), an anytime kinodynamic motion planner, that extends discontinuity-bounded CBS to rigidly-coupled systems. Our approach proposes a tri-level conflict detection and resolution framework that includes the physical coupling between the robots. Moreover, pc-dbCBS alternates iteratively between state space representations, thereby preserving probabilistic completeness and asymptotic optimality while relying only on single-robot motion primitives.

Across 25 simulated and six real-world problems involving multirotors carrying a cable-suspended payload and differential-drive robots linked by rigid rods, pc-dbCBS solves up to 92\% more instances than a state-of-the-art baseline and plans trajectories that are 50–60\% faster while reducing planning time by an order of magnitude.
\end{abstract}

\section{Introduction}
Physically-coupled systems, such as multirotors collaboratively transporting cable-suspended payloads~\cite{sreenath2013dynamics} or multiple mobile manipulators transporting objects~\cite{tallamraju2019motion}, are increasingly used in real-world tasks requiring coordinated interaction. 
These systems are particularly valuable in environments such as construction sites for carrying tools or materials and in precision tasks requiring synchronized motion.
The robot coupling introduces additional challenges, as the planned motions must respect both inter-robot dependencies and the system's dynamic constraints.
There has been a significant focus on controlling such systems \cite{sun2023nonlinear,li2023nonlinear} and only limited work on planning feasible motions in cluttered environments that require team formation changes.
Moreover, existing planners produce motions that are rather slow and fail to exploit the agility of the underlying single-robot systems. This limitation arises from the reliance on a simplified model of the system, where the planner generates suboptimal, almost quasi-static plans~\cite{wahba2024kinodynamic}. 

\begin{figure}[ht]
  \centering
    \includegraphics[width=0.23\textwidth]{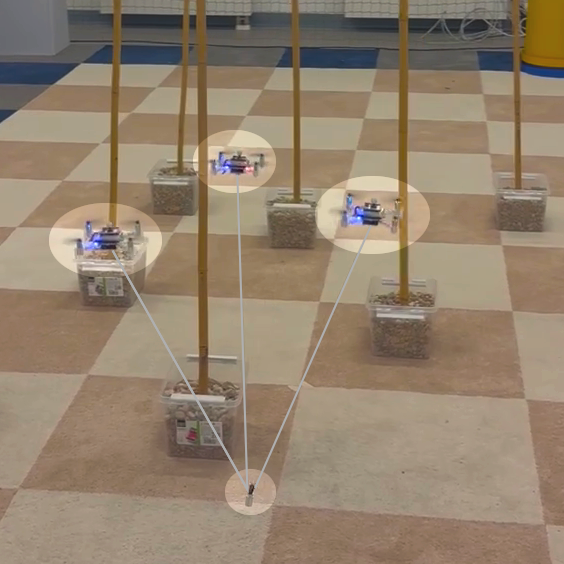}
    \includegraphics[width=0.23\textwidth]{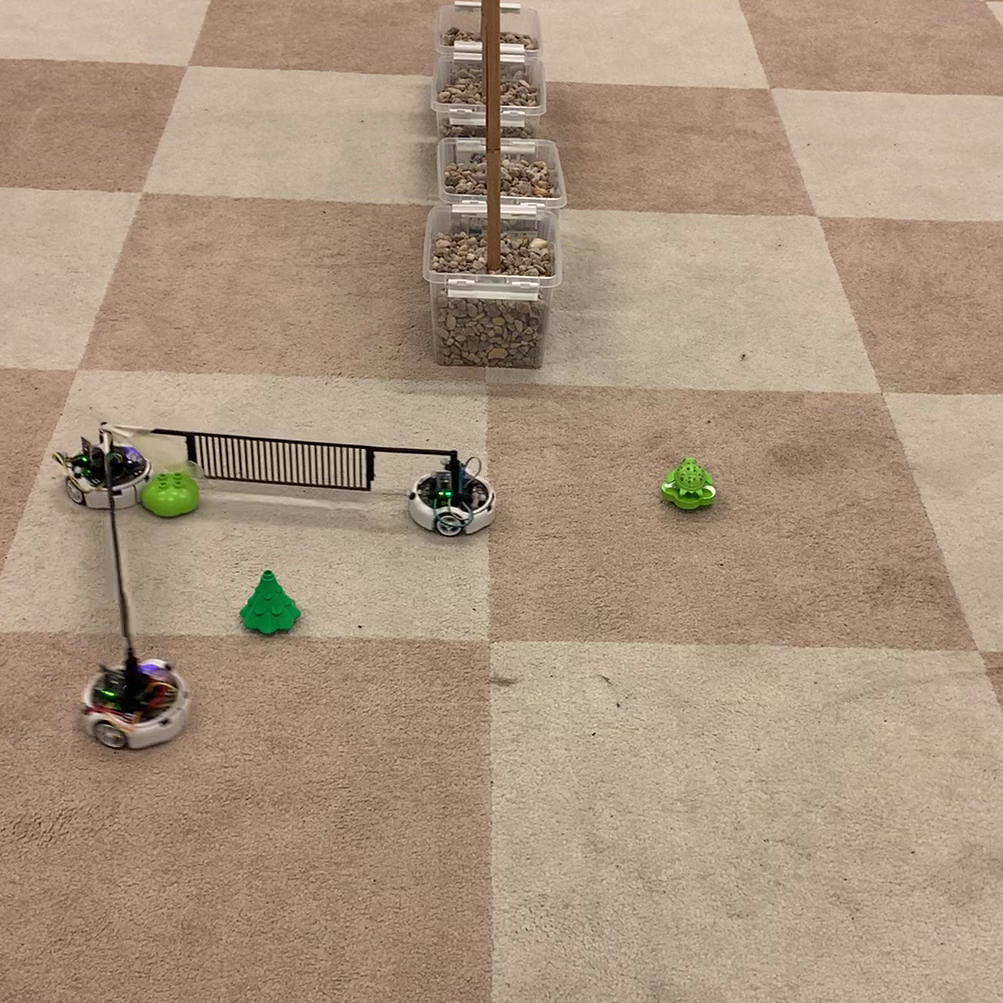}
  \caption{Real experiments validation scenarios. Left: Three multirotors transporting a cable-suspended payload in a forest-like environment. Right: Three differential drive robots connected with rigid rods collecting items while avoiding obstacles.}
  \label{fig:real_envs}
\end{figure}

In this work, we address the limitations of the current state-of-the-art~\cite{wahba2024kinodynamic},  and present an anytime, probabilistically complete, motion planner for physically-coupled multi-robot systems in cluttered environments. To the best of our knowledge, this is the first work to offer this combination of simplicity, strong theoretical guarantees, and superior performance in the physically-coupled systems domain. 
Our approach builds on discontinuity-bounded Conflict-Based Search (db-CBS)~\cite{moldagalieva2024db}, a multi-robot motion planning algorithm designed for uncoupled robots. 
Although db-CBS reasons over the stacked state space of the robots, it only accounts for collision avoidance and neglects other interactions. 
Directly enforcing physical coupling constraints during trajectory optimization is also impractical.
The  redundancy in the state space with the coupling constraints, results in numerical instabilities and ill-conditioned optimization problems, causing failure.
On the other hand, pc-dbCBS introduces a tri-level conflict detection and resolution that (i) embeds the rigidity constraints between the robots and (ii) switches between state space representations while retaining all the completeness and optimality guarantees of db-CBS.  

The main algorithmic contribution of this work is a general anytime, probabilistically complete, kinodynamic motion planning framework for \textbf{physically-coupled multi-robot systems}, relying only on single-robot motion primitives. Moreover, we provide a key idea of alternating between state space representations from the stacked to a minimal representation in the same planning framework while having completeness and asymptotic optimality guarantees. Empirically, we test our method in simulation and real experiments on two case studies: multirotors transporting payloads with cables and differential drive robots connected by rigid rods as shown in \cref{fig:real_envs}. We show that our method has higher success rate than the current state-of-the-art~\cite{wahba2024kinodynamic} and outperforms it in terms of cost, energy consumption, and computational time. 

\section{Related Work}

Existing physically-coupled multi-robot systems include humanoid robots transporting payloads~\cite{rapetti2021shared}, multi-robot mobile manipulators moving objects~\cite{tallamraju2019motion}, and aerial transport systems~\cite{afifi2022toward,wahba2024kinodynamic, gorlo2024geranos}.
The problem of aerial manipulation with a single robot~\cite{tognon2019truly} and high-dimensional articulated robots \cite{gayle2007efficient,kulkarni2020reconfigurable,saleem2021search} are also related, as they have to consider very large search spaces for motion planning.

In principle such motion planning problems can be solved with any existing single-robot kinodynamic motion planner by stacking the states and actions.
Common formulations use search-\cite{saleem2021search}, sampling-~\cite{AO-RRT}, or optimization-based~\cite{malyutaConvexOptimizationTrajectory2022a, howell2019altro} approaches.
Special extensions for the multi-robot case can improve the scalability by leveraging the sparsity of the problem.
These methods include sampling-based approaches such as K-CBS~\cite{kottinger2022conflict}, optimization and control-inspired methods such as S2M2~\cite{chen2021scalable}, or hybrid methods such as db-CBS~\cite{moldagalieva2024db}.

Motion planners for the physically-coupled case exist as well, especially for aerial transport of payloads.
One approach extends a sampling-based method to the \emph{Fly-Crane}, a system where three multirotors use two cables each to transport a rigid body~\cite{manubens2013motion}.
In cases with few or no obstacles, a formation controller might also be used, rather than a full motion planner~\cite{de2019flexible}.
Other approaches rely on differential flatness~\cite{gabellieri2023differential} that provides a simplified framework for trajectory planning by leveraging flat outputs, but it typically leads to suboptimal solutions and is unable to directly account for motor actuation limits.
Alternative methods that includes the motor actuation limits combine ideas from sampling and optimization to compute feasible trajectories~\cite{wahba2024kinodynamic,zhang2023if}.
In particular for cable-suspended payload transportation,~\cite{wahba2024kinodynamic} adopts this framework.
Here, a simplified model is first used to warm-start a trajectory optimization step that incorporates necessary constraints. However, this method has two major limitations: i) the simplified model often produces suboptimal quasi-static hovering states, and ii) the lack of feedback between levels hinders completeness by limiting solution space exploration.
One shortcoming of all existing methods is that the resulting motions are slow and do not exploit the full agile capabilities of the robots.

In contrast, this paper proposes an anytime planner that can produce time-optimal solutions, given enough computational effort. Our work is inspired by recent hybrid methods for uncoupled kinodynamic motion planning~\cite{moldagalieva2024db} and empirically compares to a hierarchical planner that was previously used for aerial transport~\cite{wahba2024kinodynamic}.

\section{Problem Formulation}

We consider the motion planning problem of $N$ robots that are physically-coupled with rigid connections, forming an actuated multi-robot system. 
The state of the system is represented as a stacked vector of each individual robot's state as follows
\begin{equation}
    \label{eq:stacked_state}
    \x = (\x^{1}, \x^{2}, \ldots, \x^{N})^\top \in \X \subseteq \Rdx,
\end{equation}
where $\x^{i} \in \X^{i} \subseteq \Rdxi$ is the state of the $i^{\text{th}}$ robot, and $\X$ is the stacked state space. Similarly, the stacked action vector is defined as
\begin{equation}
    \label{eq:stacked_action}
    \ui = (\ui^{1}, \ui^{2}, \ldots, \ui^{N})^\top \in \U \subseteq \Rdu,
\end{equation}
where $\ui^{i} \in \U^{i} \subseteq \Rdui$ is the controls applied to the $i^{\text{th}}$ robot.
The dynamics of the stacked system are governed by
\begin{equation}
    \label{eq:system_dynamics}
    \xdot = \mathbf{f}(\x, \ui).
\end{equation}

To employ gradient-based optimization, we assume that the Jacobian of $\mathbf{f}$ with respect to $\x$ and $\ui$ is available, as is typical for motion planning in actuated multi-robot systems. 
Let $\Xseq = \langle \x_0, \x_1, \hdots, \x_T \rangle$ and  $\Useq = \langle \ui_0, \ui_1, \hdots, \ui_{T-1} \rangle$  be a sequence of states and controls sampled at time $0, \Delta t, \hdots, T\Delta t$ respectively,
where $\Delta t$ is a small timestep and the controls are constant during this timestep. 
We denote the start state as $\x_s$, the goal state as $\x_f$, and the collision-free state space as $\X_{\text{free}} \subset \X$, which accounts for robots as well as collisions against the environment.
Our goal is to find a solution of states and actions from a start to a goal state in the minimal time $T$, which can be framed as the following optimization problem
\begin{align}
    &\min_{\Xseq, \Useq, T} \hspace{0.2cm} J(\Xseq, \Useq, T), \label{eq:general-optimization-problem}\\
    &\text{\noindent s.t.}\begin{cases}
     \x_{k+1} = \text{step}(\x_k, \ui_k) \quad \forall k\in\{0, \ldots, T-1\}, \nonumber \\
     \ui_k \in \mathcal{U}  \quad \forall k\in\{0, \ldots, T-1\}, \nonumber \\
     \x_0 = \x_s, \hspace{0.2cm} \x_T = \x_f, \nonumber \\
      \x_k \in \X_{\text{free}} \subset \X  \quad \forall k\in\{0, \ldots, T\}, \\
     \mathbf{g}(\x) = 0, \nonumber \\
    \end{cases}
\end{align}
where the cost function is $T$ and other task objectives (e.g., energy). 
The first constraint is the time-discretized system dynamics and the second constraint bounds actions to the admissible space $\mathcal{U}$. 
The third set of constraints enforces the given start and goal states. 
The final constraint $\mathbf{g}(\x)$, assumed to be continuously differentiable (similar to~\cite{de2005feedback}),
enforces the physical rigidity coupling between robots, ensuring their motions satisfy the kinematic and dynamic interactions. We assume that the constraint set $\{\x \in \X \mid \mathbf{g}(\x) = 0\}$ admits a local mapping $\Phi: \X_m \rightarrow \X$ with minimal coordinates $\x_m \in \X_m$. 
To demonstrate the generality of the framework across embodiments, we present two robot platforms: (i) Unicycles, low-dimensional non-holonomic robots, rigidly connected in a ``line''-formation. (ii) Multirotors transporting payloads with  cables (modeled as rigid rods), high-dimensional underactuated robots.   
These examples are illustrative only. Our framework applies to any rigidly-coupled robot teams.
\subsection{Unicycles with Rigid Rods (UR)}
Consider a team of $N$ unicycles connected by $N-1$ rigid rods of fixed lengths in a line formation. 
The state of the $i^{\text{th}}$ unicycle is described as $\x^{i} = (\px^{i}, \py^{i}, \thi^{i})^\top, $
where $\p^i = (\px^i, \py^i)^\top$ denote the position of the $\emph{i}^{th}$ unicycle, and $\thi^i$ represents its orientation. 
The kinematic model for the unicycle is given by $\xdot^{i} = \mathbf{C}^{i}\ui^{i}$ as
\begin{equation}
    \label{eq:unicycle}
    \xdot^{i} = 
    \begin{pmatrix}
        \cos(\thi^{i}) & 0 \\
        \sin(\thi^{i}) & 0 \\
        0 & 1
    \end{pmatrix}
    \begin{pmatrix}
        v^{i} \\ \omega^{i}
    \end{pmatrix},
\end{equation}
where $\ui^{i} = (v^{i}, \omega^{i})^\top$ represents the linear and angular velocities, respectively.
Let the length of the rod connecting the $i^{\text{th}}$ and $(i+1)^{\text{th}}$ unicycles be $l^i$. Then the coupling constraint $\mathbf{g}(\x)=0$ is
\begin{align}
    \| \p^{i} - \p^{i+1} \| - l^i = 0, \quad i=1,\ldots, N-1.
\end{align}

\subsection{Multirotors Transporting a Payload (MP)}
\label{sec:multirotor}

Consider a team of $N$ multirotors transporting a cable-suspended payload. 
The $i^{\text{th}}$ multirotor is modeled as a rigid floating base with state $\x^{i} = (\mathbf{p}^{i}, \mathbf{R}^{i}, \mathbf{v}^{i}, \boldsymbol{\Omega}^{i})^\top$. 
Here, $\mathbf{p}^{i}, \mathbf{v}^{i} \in \mathbb{R}^3$ represent the position and velocity in the world frame, $\mathbf{R}^{i} \in SO(3)$ represents the rotation matrix, and $\boldsymbol{\Omega}^{i} \in \mathbb{R}^3$ is the angular velocity expressed in the body frame. 
The action $\ui^{i} \in \mathbb{R}^4$ is defined as the forces at the rotors, $\ui^{i} = (f_{1}^{i}, f_{2}^{i}, f_{3}^{i}, f_{4}^{i})^\top$. 
The dynamics are derived from Newton-Euler equations for rigid bodies as
\begin{align}
    \label{eq:multirotor}
    \dot{\mathbf{p}}^{i} &= \mathbf{v}^{i}, \quad &m^i \dot{\mathbf{v}}^{i} &= \mathbf{R}^{i} f_T^{i} \ez - m^i g\ez, \quad \\
    \dot{\mathbf{R}}^{i} &= \mathbf{R}^{i} \hat{\boldsymbol{\Omega}}^{i}, \quad
    &\mathbf{J}^i \dot{\boldsymbol{\Omega}}^{i} &= \mathbf{J}^i \boldsymbol{\Omega}^{i} \times \boldsymbol{\Omega}^{i} + \mathbf{M}^{i}, \nonumber
\end{align}
where $m^i$ is the mass, $\mathbf{J}^i$ is the inertia matrix, g is the gravitational acceleration constant, $\ez = (0, 0, 1)^\top$.
The $(\hat{\cdot})$ denotes the skew-symmetric mapping $\mathbb{R}^3 \rightarrow \mathfrak{s}\mathfrak{o} (3)$. 

The collective thrust and torques, $\boldsymbol{\eta}^{i} = (f_T^{i}, \mathbf{M}^{i})^\top$, are linearly related to the motor forces $\ui^{i}$ via a fixed and known actuation matrix.

Each multirotor is connected with a cable modeled as a rigid rod of length $l^i$ to a point mass payload at $\p^0$ with mass $m^0$.
Then the coupling constraint $\mathbf{g}(\x)=0$ is
\begin{align}
    \| \p^{0} - \p^{i} \| - l^i = 0, \quad i=1,\ldots N.
\end{align}

\section{Approach}

\subsection{pc-dbCBS}

Our motion planning approach, pc-dbCBS, extends and adapts db-CBS to the high-dimensional physically-coupled systems that are subject to physical constraints between the robots. 
Building upon this foundation, pc-dbCBS utilizes the three-level iterative framework of db-CBS by integrating additional definition of conflicts and conversions of state representations to manage physical coupling constraints.
In principle, pc-dbCBS utilizes the efficiency of the discrete search with the stacked state space, allowing the usage of pre-computed motion primitives for the single robot, along with the effectiveness of trajectory optimization on a minimal representation of the coupled system. 
The pseudo code in \cref{alg:dbcbs} highlights major changes compared to db-CBS.
\subsubsection{Single Robot Planning} The \emph{first level} uses $\dbAstar$ \cite{ortiz2024idb} to plan for each robot a trajectory with state discontinuous up to $\delta$ from a graph constructed of precomputed motion primitives $\mathcal{M}$. 
A motion primitive is defined as a tuple $\langle \Xseq^i, \Useq^i, K \rangle$ of state and action sequences over $K$ steps that satisfy the dynamics of the robot. 
For each robot $i$, $\dbAstar$ outputs state and action sequences 
which adhere to the single robot dynamics, such as \cref{eq:unicycle} for unicycles and \cref{eq:multirotor} for multirotors. 
These sequences form \textit{$\delta$-discontinuity-bounded} solutions under the condition
\begin{align}
    &d(\x_{k+1}^i, \text{step}(\x_k^i, \ui_k^i)) \leq \delta \quad \forall k, \\
    &\ui_k^i \in \U^i, \quad \x_k^i \in \X^i, \nonumber \\
    &d(\x_0^i, \x_s^i) \leq \delta, \quad d(\x_K^i, \x_f^i) \leq \delta \nonumber,
\end{align}
where $d$ is a metric $d : \X^i \times \X^i \to \mathbb{R}$, which measures the distance between two states. In this work, we pre-compute each robot’s set of motion primitives offline as in db-CBS.

\begin{algorithm}[t]
    \caption{pc-dbCBS: changes to \cite{moldagalieva2024db} are highlighted}
    \label{alg:dbcbs}
    \DontPrintSemicolon
    \SetVlineSkip{2pt}

    \SetKwFunction{TerminationCriteria}{TerminationCriteria}
    \SetKwFunction{UpdateHyperParameters}{UpdateHyperParameters}
    \SetKwFunction{HLDiscreteSearch}{HLDiscreteSearch}
    \SetKwFunction{Convert}{Convert}
    \SetKwFunction{Optimization}{Optimization}
    \SetKwFunction{PriorityQueuePop}{PriorityQueuePop}
    \SetKwFunction{ComputeConflict}{ComputeConflict}
    \SetKwFunction{ExtractConstraintsAndAddToO}{ExtractConstraintsAndAddToO}
    \SetKwFunction{ResolveRobotCollision}{ResolveRobotCollision}
    \SetKwFunction{ResolvePhysicalConstraint}{ResolvePhysicalConstraint}
    \SetKwFunction{ResolvePCCollision}{ResolvePCCollision}
    \SetKwFunction{CheckForConflict}{CheckForConflict}
    \SetKwFunction{ExtractRandomConstraintAndAddToOpt}{ExtractRandomConstraintAndAddToOpt}

    \While{$\lnot$\TerminationCriteria() \label{alg:line1}}{
        $\delta, \mathcal M \leftarrow \UpdateHyperParameters()$\label{alg:line2}\;
        $X_j, U_j \leftarrow \HLDiscreteSearch(P, \delta, \mathcal M)$\label{alg:line3}\;
        \tikzmk{A}$X_d, U_d \leftarrow \Convert(X_j, U_j)$ \marklineOne{-13pt}\label{alg:line4}\;
        $X, U \leftarrow \Optimization(P, X_d, U_d)$\label{alg:line5}\;
    }

    \SetKwProg{DefOne}{def}{:}{}
    \DefOne{HLDiscreteSearch($P, \delta, \mathcal M$)}{ \label{alg:line6}
        \tcc{Initialize priority queue $\mathcal O$ by planning for individual robots independently}
        \While{$|\mathcal O| > 0$}{
            $P \leftarrow \PriorityQueuePop(\mathcal O)$ \Comment*{Lowest solution cost}
            \If{$\lnot$ \ResolveRobotCollision($P, \mathcal O$)}{ \label{alg:line9}
                \tikzmk{A}\If{$\lnot$ \ResolvePhysicalConstraint($P, \mathcal O$)}{ \label{alg:line10}
                    \If{$\lnot$ \ResolvePCCollision($P, \mathcal O$) \marklineTwo{-25pt}}{ \label{alg:line11}
                        \Return stacked solution\; \label{alg:line12}
                    }
                }
            }

        }
    }
    \SetKwProg{DefTwo}{def}{:}{}
    \DefTwo{ResolvePhysicalConstraint($P, \mathcal O$)}{ \label{alg:line13}
        \For{$k \in [1, K]$} { 
            $C \leftarrow \CheckForConflict(P.solution, k)$\;
            \If{$C \neq \emptyset$} {
                \ExtractRandomConstraintAndAddToOpt($C, \mathcal O$)\;
                return true\;
            }
        }

    }
\end{algorithm}

\subsubsection{Conflict Resolution}
The \emph{second level} employs a search to analyze the motions generated by the first level, identifying conflicts between robots, and creating constraints that are then iteratively resolved by the first level. 
Conflicts are resolved through a hierarchical process that sequentially addresses three types of conflicts:  
inter-robot collisions, physical coupling violations, and coupling elements (e.g., rods) and environment collisions. 
These conflicts are resolved in \crefrange{alg:line9}{alg:line11}.
Each type of conflict is resolved up to the tolerance $\delta$ over the full time horizon of the planned trajectory, before the next conflict type is considered.

The first level resolves robot-robot collisions by using the \texttt{ResolveRobotCollision} function (\cref{alg:line9}) to detect overlaps in planned motions. 
When conflicts are found, constraints are added to the priority queue $\mathcal{O}$, and the \texttt{HLDiscreteSearch} process iterates with these updates. This behavior is identical to db-CBS.

The second level resolves the physical coupling constraints. 
The \texttt{ResolvePhysicalConstraint} function (\cref{alg:line10}) evaluates the solution of the robots' formations up to given bounds. 
For each step $k$, the constraint $\|\mathbf{g}(\x)\| < \delta$ is checked.
If a violation is detected, only a single new node is added to $\mathcal{O}$, constraining a randomly picked single robot motion.
Note that in the traditional CBS framework $N$ nodes would need to be added, as it is unknown which combination of robots is causing the constraint violation.
Picking a single robot instead reduces the size of $\mathcal{O}$, but has some theoretical drawbacks, as discussed later.

The final level of conflict resolution involves collision checking between the physical coupling elements (e.g., rods) and the obstacles in the environment. 
The \texttt{ResolvePCCollision} function (\cref{alg:line11}) constructs artificial collision shapes for the coupling elements and evaluates whether they collide with obstacles. 
If a collision is identified, one additional node is added to $\mathcal{O}$, constraining the affected robot.
In principle, inter-cable collisions could be defined as conflicts; however, this would significantly increase computational overhead as they are already addressed during the optimization step as collision shapes (e.g., rods or cylinders), we find that this is not needed in practice.

If no conflicts are detected at all levels, the algorithm returns a stacked space solution that adheres to all robot-robot, physical coupling, and obstacle collisions constraints up to the bounded tolerance $\delta$.

\subsubsection{Stacked to Constrained Systems}
The next step involves converting the stacked space solution from the discrete search to the constrained physically-coupled dynamical system (\cref{alg:line4}).
This is achieved by representing the state into a minimal representation $\x_m \in \X_m$. 
The dynamics of the physically coupled system can be projected on the coupling constraints $\mathbf{g}(\x)$ in \cref{eq:general-optimization-problem} using the minimal representation, which are governed by $\xdot_{m} = \mathbf{f}_m(\x_{m}, \ui)$.

\subsubsection{Trajectory Optimization} 
After the discrete stacked state solution is mapped to the minimal state representation, we use trajectory optimization to refine the solution from the discrete search and to repair all the discontinuities of $\delta$.
The optimization problem is reformulated using the minimal state representation as 
\begin{align} 
    \label{eq:opticost}
    &\min_{\Xseq, \Useq, \Delta t} \hspace{0.2cm} \sum_{k} (\Delta t - \Delta t_0)^2  + \beta_1 \|\ui_k\|^2 \\ & + \beta_2 \|\ddot{\boldsymbol{x}}_m(\x_{m_k}, \ui_k)\|^2  \nonumber \\
&\text{\noindent s.t.}\begin{cases}
     \x_{m_{k+1}} = \text{step}(\x_{m_k}, \ui_k) \quad \forall k\in\{0, \ldots, T-1\}, \\
     \ui_k \in \mathcal{U}  \quad \forall k\in\{0, \ldots, T-1\},  \\
     \x_{m_0} = \x_{m_s}, \hspace{0.2cm} \x_{m_T} = \x_{m_f}, \nonumber \\
      \x_{m_k} \in \X_{m_\text{free}} \subset \X_m  \quad \forall k\in\{0, \ldots, T\}. 
    \end{cases} 
\end{align}
Here, the cost function minimizes the deviation of the time step $\Delta t$ from a nominal value $\Delta t_0$, penalizes the control effort $\|\ui_k\|^2$, and reduces the system's dynamic accelerations $\|\ddot{\boldsymbol{x}}(\x_k, \ui_k)\|^2$ to improve trajectory smoothness. 
The weighting parameters $\beta_1$ and $\beta_2$ are used to balance the contributions of the control effort and dynamic smoothness terms.
This part is identical to prior formulations such as~\cite{wahba2024kinodynamic}.

\subsubsection{Anytime Planning}
\label{sec:anytimeplanning}
Our method iteratively refines solutions by updating two parameters (\cref{alg:line2}). 
First, $\delta$ is gradually decreased with a predefined rate.
Second, solutions from the optimization step, including failed ones, are extracted, split, and transformed from the minimal representation to single-robot motion primitives. 
Then these are added to the motion primitive database $\mathcal{M}$ alongside newly sampled primitives with a predefined rate. 

\subsection{Case Study: Unicycles with Rigid Rods (UR)}

\subsubsection{Constrained Dynamics}
The system can be minimally represented by the state vector
\begin{equation}
    \label{eq:unicyclemin}
    \x_{m} = (\px^{1}, \py^{1}, \thi^{1}, \cdots, \thi^{N}, \alpha^{1}, \cdots, \alpha^{N-1})^\top,
\end{equation}
where $\alpha^{i}$ represents the orientation of the $i^{\text{th}}$ rod.
We have $\x_{m} \in \mathbb R^{2N} \times (SO(2))^{2N-1}$, and the position of the $({i+1})^{\text{th}}$ unicycle is expressed by 
\begin{equation}
    \label{eq:unicycles_constraints}
    \px^{i+1} = \px^{i} + l^i\cos(\alpha^{i}), \quad 
    \py^{i+1} = \py^{i} + l^i\sin(\alpha^{i}). 
\end{equation}

The kinematics of the $N$ unicycles with rigid rods are governed by the constraints imposed by the distance between the connected unicycles.
Let the relative position and its derivative with respect to time of two neighboring robots $i$ and $j$ be
\begin{align}
    \label{eq:constraint1}
    dx_{ij} &= \px^{j} - \px^{i}, & dy_{ij} &= \py^{j} - \py^{i} \\
 \dot{dx}_{ij} &= \dot{p}_x^{j} - \dot{p}_x^{i}, & \dot{dy}_{i} &= \dot{p}_y^{j} - \dot{p}_y^{i} \nonumber
\end{align}
Thus, the constraint enforcing a fixed distance between the $i^{\text{th}}$ and $j^{\text{th}}$ consecutive unicycles is given by
\begin{equation}
    \label{eq:constraint}
    dx_{ij}^2 + dy_{ij}^2 = {l^i}^2.
\end{equation}
Differentiating \cref{eq:constraint} with respect to time yields
\begin{equation}
    \label{eq:velocity_constraint}
    2dx_{ij} \dot{dx}_{ij} + 2dy_{ij} \dot{dy}_{ij} = 0,
\end{equation}
which ensures that the velocities of the unicycles are consistent with the physical constraints imposed by the rods.
To incorporate these constraints into the system kinematics, we construct the Jacobian matrix \(\mathbf{A}\) with dimensions $(n-1) \times 3n$.
The rod constraints in \cref{eq:velocity_constraint} can then be expressed compactly as
\begin{equation}
    \mathbf{A} \xdot^{ur} = 0,
\end{equation}
where $\x^{ur} = (\px^{1}, \py^{1}, \thi^{1}, \cdots,\px^{N}, \py^{N}, \thi^{N})$.
This formulation ensures that the system dynamics remain consistent with the physical constraints imposed by the rods.

Inspired by \cite{de2005feedback}, the kinematics are then projected into the constraint-consistent space using a projection matrix $\mathbf{G}\in \mathbb R^{3n\times2n}$
\begin{align}
    \label{eq:urdyn}
    \dot{\x}^{ur} = \mathbf{G} \mathbf{u}, &\quad \mathbf{G} = \mathbf{B} - \mathbf{A}^\dagger (\mathbf{A} \mathbf{B}),
\end{align}
with $\mathbf{A}^\dagger = \mathbf{A}^\top (\mathbf{A} \mathbf{A}^\top)^{-1}$. The input mapping matrix for the unicycles $\mathbf{B} \in \mathbb R^{2n \times 2n}$ is a block-diagonal matrix defined as $\mathbf{B} = \text{diag}(\mathbf{C}^{1}, \mathbf{C}^{2}, \dots, \mathbf{C}^{N}),$
where each $\mathbf{C}^{i}$ is defined in \eqref{eq:unicycle}.

The angle of each rod $\alpha^{i}$ and its angular velocity $\dot{\alpha}^{i}$ are computed by
\begin{align}
    \label{eq:rodangle}
\alpha^{i} &= \arctan(dy_{ij}, dx_{ij})  \\ 
\quad \dot{\alpha}^{i} &= \frac{1}{{l^i}^2}dx_{ij}\dot{d}y_{ij} - dy_{ij}\dot{d}x_{ij} \quad, i \in \{1, \ldots, N\}\nonumber.
\end{align}
The dynamics $\mathbf{f}_m$ is then computed using \eqref{eq:urdyn}, \eqref{eq:rodangle} and the action vector $\ui$.

\subsubsection{Conflicts}

The conflict in the second level arises from maintaining a certain distance decided by the length of the rigid rod.
Consider $\lcki = \sqrt{dx_{ij,k}^2 + dy_{ij,k}^2} \quad \forall (i,i+1)$ to be the actual relative distance between each consecutive pair of unicycles at the $k^{\text{th}}$ step.
Then, a conflict for the robot pair $(i, i+1)$ occurs if
\begin{equation}
    | \lcki - l^i | > \delta,
\end{equation}
where $l^i$ is the nominal rod length, and $\delta$ defines the allowable tolerance.

\begin{figure*}[ht]
  \centering
    \includegraphics[width=0.23\textwidth]{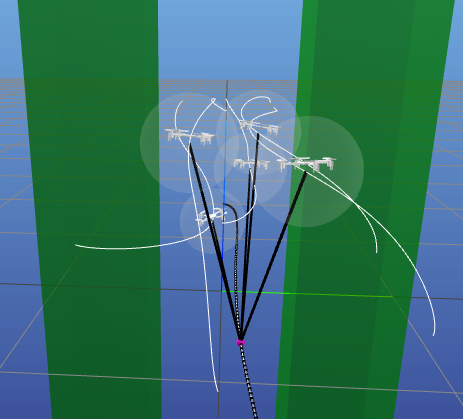}
    \includegraphics[width=0.23\textwidth]{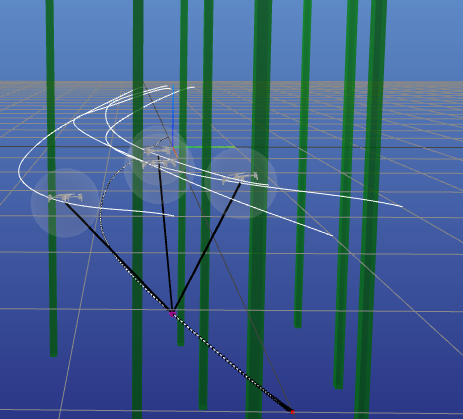}
    \includegraphics[width=0.23\textwidth]{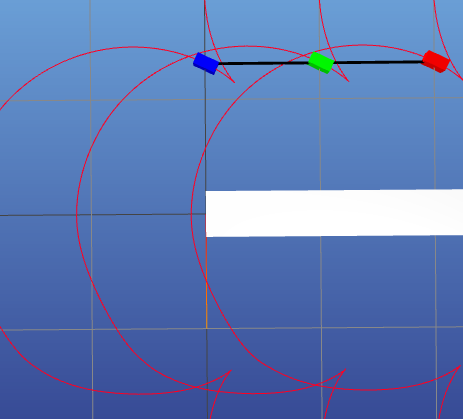}
    \includegraphics[width=0.23\textwidth]{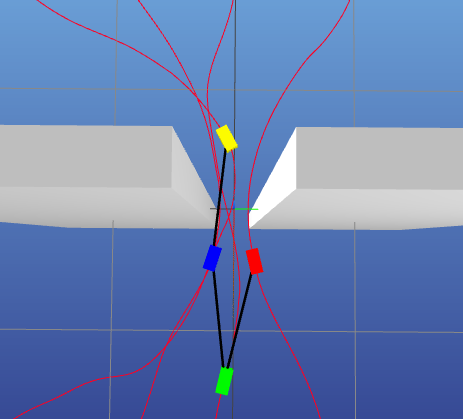}
  \caption{Simulation environments from left to right: window (5 multirotors), forest (4 multirotors), wall (3 unicycles), window (4 unicycles). Note that the forest environment is the same for both robot types. }
  \label{fig:sim_envs}
\end{figure*}
\subsection{Case Study: Multirotors Transporting a Payload (MP)}

\subsubsection{Constrained Dynamics}

The dynamics of multirotors with cable-suspended payloads, as presented in~\cite{wahba2024kinodynamic}, are described using the minimal state representation
\begin{equation}
    \label{eq:statespace}
    \x_m = (\po, \vo, \scfirst, \wifirst, \mathbf{R}^1, \wuavfirst, \ldots, \scn, \win, \mathbf{R}^N, \wuavn)^\top,
\end{equation} 
where $\mathbf{p}^0\in \mathbb{R}^3$ is the payload's position, $\vo \in \mathbb{R}^3$ is the payload velocity, $\sci \in \mathbb{S}^2$ are the cable unit vectors pointing from the UAV to the payload (with $\mathbb{S}^2 = \{\mathbf{s} \in \mathbb{R}^3 \big| \|\mathbf{s}\| = 1\}$), and $\boldsymbol{\omega}^i \in \mathbb{R}^3$ are the cable angular velocities, where $i \in \{1, \ldots, N\}$. 

The UAV position and velocity vectors are $\p^i\in \mathbb{R}^3$ and $\dot{\p}^i \in \mathbb{R}^3$ are computed as 
\begin{equation}
    \label{eq:uavpos}
    \p^i = \po - l^i\sci, \quad \dot\p^i = \dot \p^0 - l^i \scidot.
\end{equation}
 
The dynamics $\mathbf{f}_m$ of the system is defined as
\begin{align}
    \label{eq:mpdynamics}
    &\scidot = \wi \times \sci, \quad \sci = \frac{\p^0 - \p^i}{\|\p^0 - \p^i\|},\\
    &\mathbf{M_{t}}(\podd +g\ez)  = \sum_{i=1}^n (f_{T}^i\R^i\ez - m^i l^i\|\wi\|^2\sci), \nonumber\\
    &m^i l^i\dot{\boldsymbol{\omega}}^i = m^i \scihat(\podd +g\ez) - f^i_{T}\scihat\R^i\ez, \nonumber\\
    &\dot{\mathbf{R}}^i = \mathbf{R}^i \hat{\wuav}^i,
    \quad \mathbf{J} \wdot^i = \mathbf{J} \wuav^i \times \wuav^i + \mathbf{M}^i, \nonumber
\end{align}
where $\podd$ is the payload acceleration, $\mathbf{M}_{t} = m^0 \mathbf{I}_{3} + \sum_{i=1}^n m^i \sci{\sci}^\top$ and $g$ is the gravitational acceleration.

\subsubsection{Conflicts}

\renewcommand{\arraystretch}{1.02} %
\setlength{\tabcolsep}{6pt}       %
\begin{table*}[ht]
\caption{Simulation Results for multirotors with payload (MP) and unicycles with rods (UR).
Shown are mean values for the success rate, cost and computational time over 10 trials with a time limit of \SI{350}{s}. Gray: standard deviation. F: failed.}
\centering
\footnotesize
\begin{tabular}{|c||c|c|c|c||c|c|c|c||c|c|c|c|}
\hline
\multirow{3}{*}{\textbf{Environment}}
& \multicolumn{4}{c||}{\textbf{Success} [\%] $\uparrow$}
& \multicolumn{4}{c||}{\textbf{Cost} [s] $\downarrow$}
& \multicolumn{4}{c|}{\textbf{Time} [s] $\downarrow$} \\
\cline{2-13}
& \multicolumn{2}{c|}{\textbf{UR}} & \multicolumn{2}{c||}{\textbf{MP}}
& \multicolumn{2}{c|}{\textbf{UR}} & \multicolumn{2}{c||}{\textbf{MP}}
& \multicolumn{2}{c|}{\textbf{UR}} & \multicolumn{2}{c|}{\textbf{MP}} \\
\cline{2-13}
& \scriptsize \textbf{Ours} & \scriptsize \textbf{BL} & \scriptsize \textbf{Ours} & \scriptsize \textbf{BL}
& \scriptsize \textbf{Ours} & \scriptsize \textbf{BL} & \scriptsize \textbf{Ours} & \scriptsize \textbf{BL}
& \scriptsize \textbf{Ours} & \scriptsize \textbf{BL} & \scriptsize \textbf{Ours} & \scriptsize \textbf{BL} \\
\hline
Window, 2 robots
&
\scriptsize
\textbf{100.0}
&
\scriptsize
\textbf{100.0}
&
\scriptsize
\textbf{{100.0}}
&
\scriptsize
90.0
&
\scriptsize
{\textbf{{4.9}}\hspace{0.5em}{\tiny \textcolor{gray}{0.0}}}
&
\scriptsize
10.4 {\tiny \textcolor{gray}{0.1}}
&
\scriptsize
{\textbf{{2.0}}\hspace{0.5em}{\tiny \textcolor{gray}{0.1}}}
&
\scriptsize
5.5 {\tiny \textcolor{gray}{0.9}}
&
\scriptsize
{\textbf{{0.2}}\hspace{0.5em}{\tiny \textcolor{gray}{0.1}}}
&
\scriptsize
350.7 {\tiny \textcolor{gray}{0.1}}
&
\scriptsize
{\textbf{{5.2}}\hspace{0.5em}{\tiny \textcolor{gray}{1.4}}}
&
\scriptsize
364.8 {\tiny \textcolor{gray}{2.6}}
\\
Window, 3 robots
&
\scriptsize
\textbf{100.0}
&
\scriptsize
\textbf{100.0}
&
\scriptsize
\textbf{{100.0}}
&
\scriptsize
70.0
&
\scriptsize
{\textbf{{6.1}}\hspace{0.5em}{\tiny \textcolor{gray}{0.4}}}
&
\scriptsize
14.9 {\tiny \textcolor{gray}{1.6}}
&
\scriptsize
{\textbf{{2.0}}\hspace{0.5em}{\tiny \textcolor{gray}{0.2}}}
&
\scriptsize
5.2 {\tiny \textcolor{gray}{0.1}}
&
\scriptsize
{\textbf{{2.1}}\hspace{0.5em}{\tiny \textcolor{gray}{2.2}}}
&
\scriptsize
351.5 {\tiny \textcolor{gray}{0.2}}
&
\scriptsize
{\textbf{{14.0}}\hspace{0.5em}{\tiny \textcolor{gray}{5.5}}}
&
\scriptsize
376.0 {\tiny \textcolor{gray}{4.1}}
\\
Window, 4 robots
&
\scriptsize
80.0
&
\scriptsize
\textbf{{90.0}}
&
\scriptsize
\textbf{{100.0}}
&
\scriptsize
0.0
&
\scriptsize
{\textbf{{9.1}}\hspace{0.5em}{\tiny \textcolor{gray}{0.4}}}
&
\scriptsize
15.4 {\tiny \textcolor{gray}{3.1}}
&
\scriptsize
{\textbf{{2.1}}\hspace{0.5em}{\tiny \textcolor{gray}{0.4}}}
&
\scriptsize
F
&
\scriptsize
{\textbf{{54.5}}\hspace{0.5em}{\tiny \textcolor{gray}{78.6}}}
&
\scriptsize
352.7 {\tiny \textcolor{gray}{1.0}}
&
\scriptsize
{\textbf{{41.4}}\hspace{0.5em}{\tiny \textcolor{gray}{18.5}}}
&
\scriptsize
F
\\
Window, 5 robots
&
\scriptsize
90.0
&
\scriptsize
\textbf{{100.0}}
&
\scriptsize
\textbf{{100.0}}
&
\scriptsize
10.0
&
\scriptsize
{\textbf{{7.7}}\hspace{0.5em}{\tiny \textcolor{gray}{0.2}}}
&
\scriptsize
13.3 {\tiny \textcolor{gray}{1.8}}
&
\scriptsize
{\textbf{{2.1}}\hspace{0.5em}{\tiny \textcolor{gray}{0.2}}}
&
\scriptsize
5.1 {\tiny \textcolor{gray}{0.0}}
&
\scriptsize
{\textbf{{36.3}}\hspace{0.5em}{\tiny \textcolor{gray}{39.5}}}
&
\scriptsize
353.1 {\tiny \textcolor{gray}{0.6}}
&
\scriptsize
{\textbf{{90.4}}\hspace{0.5em}{\tiny \textcolor{gray}{34.2}}}
&
\scriptsize
433.3 {\tiny \textcolor{gray}{0.0}}
\\
Window, 6 robots
&
\scriptsize
\textbf{{60.0}}
&
\scriptsize
50.0
&
\scriptsize
30.0
&
\scriptsize
30.0
&
\scriptsize
{\textbf{{10.4}}\hspace{0.5em}{\tiny \textcolor{gray}{0.9}}}
&
\scriptsize
19.2 {\tiny \textcolor{gray}{2.5}}
&
\scriptsize
{\textbf{{2.8}}\hspace{0.5em}{\tiny \textcolor{gray}{0.6}}}
&
\scriptsize
5.7 {\tiny \textcolor{gray}{0.4}}
&
\scriptsize
{\textbf{{24.5}}\hspace{0.5em}{\tiny \textcolor{gray}{13.6}}}
&
\scriptsize
356.5 {\tiny \textcolor{gray}{1.4}}
&
\scriptsize
{\textbf{{153.4}}\hspace{0.5em}{\tiny \textcolor{gray}{46.9}}}
&
\scriptsize
511.2 {\tiny \textcolor{gray}{42.2}}
\\
\hline
Forest, 2 robots
&
\scriptsize
\textbf{100.0}
&
\scriptsize
\textbf{100.0}
&
\scriptsize
\textbf{100.0}
&
\scriptsize
\textbf{100.0}
&
\scriptsize
{\textbf{{8.8}}\hspace{0.5em}{\tiny \textcolor{gray}{0.0}}}
&
\scriptsize
12.5 {\tiny \textcolor{gray}{1.1}}
&
\scriptsize
{\textbf{{2.2}}\hspace{0.5em}{\tiny \textcolor{gray}{0.2}}}
&
\scriptsize
6.2 {\tiny \textcolor{gray}{1.3}}
&
\scriptsize
{\textbf{{0.9}}\hspace{0.5em}{\tiny \textcolor{gray}{0.8}}}
&
\scriptsize
350.9 {\tiny \textcolor{gray}{0.2}}
&
\scriptsize
{\textbf{{22.6}}\hspace{0.5em}{\tiny \textcolor{gray}{29.2}}}
&
\scriptsize
364.6 {\tiny \textcolor{gray}{1.7}}
\\
Forest, 3 robots
&
\scriptsize
\textbf{{100.0}}
&
\scriptsize
90.0
&
\scriptsize
80.0
&
\scriptsize
\textbf{{90.0}}
&
\scriptsize
{\textbf{{10.8}}\hspace{0.5em}{\tiny \textcolor{gray}{0.1}}}
&
\scriptsize
12.7 {\tiny \textcolor{gray}{0.5}}
&
\scriptsize
{\textbf{{2.5}}\hspace{0.5em}{\tiny \textcolor{gray}{0.5}}}
&
\scriptsize
5.3 {\tiny \textcolor{gray}{0.1}}
&
\scriptsize
{\textbf{{4.3}}\hspace{0.5em}{\tiny \textcolor{gray}{10.2}}}
&
\scriptsize
351.1 {\tiny \textcolor{gray}{0.3}}
&
\scriptsize
{\textbf{{86.3}}\hspace{0.5em}{\tiny \textcolor{gray}{38.1}}}
&
\scriptsize
375.8 {\tiny \textcolor{gray}{6.2}}
\\
Forest, 4 robots
&
\scriptsize
\textbf{{100.0}}
&
\scriptsize
50.0
&
\scriptsize
70.0
&
\scriptsize
\textbf{{90.0}}
&
\scriptsize
{\textbf{{11.8}}\hspace{0.5em}{\tiny \textcolor{gray}{0.1}}}
&
\scriptsize
15.9 {\tiny \textcolor{gray}{0.8}}
&
\scriptsize
{\textbf{{2.5}}\hspace{0.5em}{\tiny \textcolor{gray}{0.1}}}
&
\scriptsize
5.3 {\tiny \textcolor{gray}{0.1}}
&
\scriptsize
{\textbf{{5.8}}\hspace{0.5em}{\tiny \textcolor{gray}{7.7}}}
&
\scriptsize
353.3 {\tiny \textcolor{gray}{0.5}}
&
\scriptsize
{\textbf{{64.3}}\hspace{0.5em}{\tiny \textcolor{gray}{32.3}}}
&
\scriptsize
380.9 {\tiny \textcolor{gray}{12.0}}
\\
Forest, 5 robots
&
\scriptsize
\textbf{{90.0}}
&
\scriptsize
30.0
&
\scriptsize
0.0
&
\scriptsize
\textbf{{80.0}}
&
\scriptsize
{\textbf{{14.3}}\hspace{0.5em}{\tiny \textcolor{gray}{0.2}}}
&
\scriptsize
19.8 {\tiny \textcolor{gray}{2.8}}
&
\scriptsize
F
&
\scriptsize
{\textbf{{6.9}}\hspace{0.5em}{\tiny \textcolor{gray}{0.3}}}
&
\scriptsize
{\textbf{{10.4}}\hspace{0.5em}{\tiny \textcolor{gray}{11.1}}}
&
\scriptsize
355.0 {\tiny \textcolor{gray}{0.9}}
&
\scriptsize
F
&
\scriptsize
{\textbf{{485.6}}\hspace{0.5em}{\tiny \textcolor{gray}{63.3}}}
\\
Forest, 6 robots
&
\scriptsize
\textbf{{60.0}}
&
\scriptsize
0.0
&
\scriptsize
10.0
&
\scriptsize
\textbf{{70.0}}
&
\scriptsize
{\textbf{{16.1}}\hspace{0.5em}{\tiny \textcolor{gray}{0.1}}}
&
\scriptsize
F
&
\scriptsize
{\textbf{{3.2}}\hspace{0.5em}{\tiny \textcolor{gray}{0.0}}}
&
\scriptsize
6.5 {\tiny \textcolor{gray}{0.8}}
&
\scriptsize
{\textbf{{57.4}}\hspace{0.5em}{\tiny \textcolor{gray}{74.6}}}
&
\scriptsize
F
&
\scriptsize
{\textbf{{227.3}}\hspace{0.5em}{\tiny \textcolor{gray}{0.0}}}
&
\scriptsize
478.7 {\tiny \textcolor{gray}{19.1}}
\\
\hline
\end{tabular}
\label{table1}
\end{table*}

A challenge for the conflict detection is that the constraint $\mathbf{g}(\x)$ depends on the payload position $\p^0$, which is not part of the stacked state.
Thus, we estimate the payload positions $\poest_k$ by solving the following optimization problem
\begin{align}
    \label{eq:payload-optimization}
    \min_{\poest_k} \sum_{i=1}^{N} &(\|\poest_k - \mathbf{p}_k^i\| - l^i)^2 + 
    \mu \|\poest_k - \p^{0_d}_{k}\| \\ & +  
    \lambda (\min_{i}\{p_{z_k}^i\} - p^{0_e}_{z_k} - l_{\text{min}})^2,    \nonumber
\end{align}
where $\mu$ and $\lambda$ are weighting parameters, $\p^{0_d}_{k}$ is used to penalize jumps of the payload estimate from the previous solution, and $l_{\text{min}}$ is the cable length associated with the robot closest to the payload $\min_{i}\{p_{z_k}^i\}$ in the $z$-axis among all robots. 
This cost function ensures that the payload position minimizes deviations from nominal cable lengths, penalizes significant changes in the payload position, and guarantees the estimated payload position remains below the robots' positions.

Once the payload position $\poest_k$ is estimated, the actual cable lengths are computed as
\begin{equation}
    \lcki = \|\poest_k - \mathbf{p}_k^i\|, \quad \forall i \in \{1, \dots, N\}.
\end{equation}
A conflict in the second level for all robots arises if any cable length $\lcki$ deviates from the nominal cable length $l^i$ by more than the tolerance $\delta$
\begin{equation}
    | \lcki - l^i | > \delta.
\end{equation}
Note that in this case all robots participate in the conflict.

\subsection{Theoretical Remarks}

pc-dbCBS inherits the probabilistic completeness and asymptotic optimality of db-CBS~\cite[Theorem 1]{moldagalieva2024db}. 
In particular \cite[Theorem 1]{moldagalieva2024db} states that db-CBS is asymptotically optimal (implying probabilistic completeness), i.e.,
\begin{equation} \label{dbcbs-proof}
    \lim_{n\to\infty}P(c_n - c^{\ast} > \varepsilon)=0\quad\forall\,\varepsilon>0,
\end{equation}
where $c_n$ is the cost in iteration $n$ and $c^{\ast}$ is the optimal cost. In db-CBS, at each iteration $n$, the discrete search finds the optimal solution within $\delta$, if one exists, yielding the optimal discrete cost~$c_n$. 
At each iteration the motion primitive library grows and $\delta$ shrinks, expanding the discrete search graph. 
For pc-dbCBS, the same argument of the proof as in~\cite[Theorem 1]{moldagalieva2024db} still holds with the highlighted changes in~\cref{alg:dbcbs}, because the outer-loop (\crefrange{alg:line1}{alg:line5}) is unchanged.

We note that the inner loop \crefrange{alg:line6}{alg:line12} of our proposed algorithm violates some key assumptions of CBS: i) for completeness, all possible alternatives need to be considered in the open list (which we violate in \texttt{ResolvePhysicalConstraints} by only including a randomly-picked single new entry); and ii) for optimality, we need to resolve conflicts in the order of their occurrence in time (which we violate using our hierarchical approach that checks for a certain type of conflict over the whole time horizon).
These changes might result producing near-optimal results of a single inner loop iteration; however the discrete search does not deterministically prune potential solution trajectories, thus the key properties of asymptotic optimality and probabilistic completeness are retained.
However, our choice impacts the runtime as potentially more outer loop iterations are needed.

\section{Results}

To validate the performance of our method, we compare with a state-of-the-art baseline method for physically-coupled multi-robot kinodynamic planning~\cite{wahba2024kinodynamic}, in both simulation and real experiments. 
We evaluate both methods on two systems: unicycles connected by rigid rods and multirotors transporting payloads with cables.
To this end, we extend the baseline to the unicycle case with rigid rods.

The baseline relies in the first planning stage on OMPL~\cite{OMPL} and for optimization on Dynoplan~\cite{ortiz2024idb}, a motion planning framework built on Crocoddyl~\cite{Crocoddyl}. 
We extend the implementation of Dynoplan to include the unicycles with rods, implemented in C++. Both kinodynamic motion planners use the Flexible Collision Library (FCL)~\cite{FCL} for collision checking.
For solving \eqref{eq:payload-optimization}, we rely on NLopt.
We will publicly release the code and problem instances after the double-blind peer-review.

\subsection{Simulations}

For multirotors with cable-suspended payload, we validate our method in simulation using a software-in-the-loop setup. 
This setup executes the actual flight controller code~\cite{wahba2023efficient} designed for Bitcraze Crazyflie 2.1 multirotors. 
For unicycles, we implemented a nonlinear controller~\cite{kanayama1990stable} in Python.
In simulation, as shown in \cref{fig:sim_envs}, we test pc-dbCBS (\textbf{Ours}) and the baseline (\textbf{BL}) on three distinct scenarios for both systems (see supplemental video). 
For $n$-unicycles with rods and multirotors with cable-suspended payloads, the first scenario involves a window environment where the goal is to pass through a narrow passage.
The second scenario is a forest-like environment with dense obstacles. 
To evaluate the completeness of both methods, we designed a third scenario specifically for unicycles with rods:  we constrain the angular velocity of each unicycle in a wall environment, allowing only clockwise rotations $(\omega^i \in [0, 0.5] \ \si{rad/s})$.
For each scenario, we evaluate five different problem instances, gradually increasing the number of robots from two to six. 
We denote (F) as failed attempts.

\renewcommand{\arraystretch}{1.02} %
\setlength{\tabcolsep}{6pt}       %
\begin{table}[t]
\caption{Simulation Results for unicycles with rods (UR).
Shown are mean values for the success rate, cost, and computational time over 10 trials with a time limit of \SI{350}{s}.
}
\centering
\footnotesize
\begin{tabular}{|c||c|c||c|c||c|c|}
\hline
\multirow{3}{*}{\textbf{Environment}}
& \multicolumn{2}{c||}{\textbf{Success [\%]} $\uparrow$}
& \multicolumn{2}{c||}{\textbf{Cost} [s] $\downarrow$}
& \multicolumn{2}{c|}{\textbf{Time} [s] $\downarrow$} \\
\cline{2-7}
& \scriptsize \textbf{Ours} & \scriptsize \textbf{BL}
& \scriptsize \textbf{Ours} & \scriptsize \textbf{BL}
& \scriptsize \textbf{Ours} & \scriptsize \textbf{BL} \\
\cline{2-7}
\hline
Wall, 2 robots
&
\scriptsize
\textbf{{90.0}}
&
\scriptsize
0.0
&
\scriptsize
{\textbf{{13.0}}\hspace{0.5em}{\tiny \textcolor{gray}{0.6}}}
&
\scriptsize
F
&
\scriptsize
{\textbf{{3.7}}\hspace{0.5em}{\tiny \textcolor{gray}{5.1}}}
&
\scriptsize
F
\\
Wall, 3 robots
&
\scriptsize
\textbf{{90.0}}
&
\scriptsize
0.0
&
\scriptsize
{\textbf{{12.7}}\hspace{0.5em}{\tiny \textcolor{gray}{0.2}}}
&
\scriptsize
F
&
\scriptsize
{\textbf{{18.0}}\hspace{0.5em}{\tiny \textcolor{gray}{14.7}}}
&
\scriptsize
F
\\
Wall, 4 robots
&
\scriptsize
\textbf{{50.0}}
&
\scriptsize
0.0
&
\scriptsize
{\textbf{{12.6}}\hspace{0.5em}{\tiny \textcolor{gray}{0.0}}}
&
\scriptsize
F
&
\scriptsize
{\textbf{{18.5}}\hspace{0.5em}{\tiny \textcolor{gray}{24.4}}}
&
\scriptsize
F
\\
Wall, 5 robots
&
\scriptsize
\textbf{{80.0}}
&
\scriptsize
0.0
&
\scriptsize
{\textbf{{12.8}}\hspace{0.5em}{\tiny \textcolor{gray}{0.3}}}
&
\scriptsize
F
&
\scriptsize
{\textbf{{56.8}}\hspace{0.5em}{\tiny \textcolor{gray}{55.2}}}
&
\scriptsize
F
\\
Wall, 6 robots
&
\scriptsize
\textbf{{70.0}}
&
\scriptsize
0.0
&
\scriptsize
{\textbf{{12.9}}\hspace{0.5em}{\tiny \textcolor{gray}{0.4}}}
&
\scriptsize
F
&
\scriptsize
{\textbf{{25.3}}\hspace{0.5em}{\tiny \textcolor{gray}{22.0}}}
&
\scriptsize
F
\\
\hline
\end{tabular}
\label{table2}
\end{table}

All experiments were conducted on a workstation (AMD Ryzen Threadripper PRO 5975WX @ 3.6 GHz, 128 GB RAM, Ubuntu 22.04). 
Each experiment was repeated 10 times, with a runtime limit for both planners of \SI{350}{s}.
\subsubsection{Comparison Metrics}

To evaluate the performance of our method, we consider three key metrics on 25 problem instances: success rate, cost, and computation time. The results are summarized in Tables \ref{table1} and \ref{table2}.
The success rate is defined as the proportion of problem instances where a feasible trajectory is successfully computed by the motion planner.
The cost metric is the total time of the trajectory from the optimization process for both methods, as both utilize the same trajectory optimization step. 
Finally, the computation time is the motion planner's runtime required to compute a feasible trajectory.

\subsubsection{Results}

For multirotors with cable-suspended payloads tested in window and forest environments, pc-dbCBS consistently generates $60\%$ lower cost (i.e., faster trajectories) than the baseline. 
\textbf{Ours} achieves a higher success rate in narrow-passage scenarios(e.g., window environment).

For unicycles connected by rods in dense environments, pc-dbCBS outperforms the baseline in both success rate and cost. 
However, in dense scenes like the forest for multirotors with cable-suspended payloads, the success rate of pc-dbCBS decreases as the team grows, due to the combinatorial explosion in cable-obstacles collision conflicts, as the finite set of motion primitives cannot explore every branch in the graph. These failures arise from practical limits on time and the number of primitives, thus, this does not contradict with pc-dbCBS retaining its probabilistic completeness. 

Furthermore, pc-dbCBS demonstrates in \cref{table2} high success rates in the wall environment, where unicycles are restricted to clockwise rotation. 
In contrast, the baseline fails to generate any feasible trajectories in this scenario, because it does not reason about dynamic limits, causing the optimization step to consistently fail.

In terms of computation time, the geometric planner in the baseline executes until it times out, followed by additional time for the optimization step. 
In contrast, pc-dbCBS applies the anytime planning property across the entire framework, allowing us to report both the cost and computation time of the first valid solution, highlighting its efficiency in finding solutions fast with better cost.

\subsubsection{Optimality and Anytime Planning}
\begin{figure}
    \centering
    \includegraphics[width=0.9\linewidth]{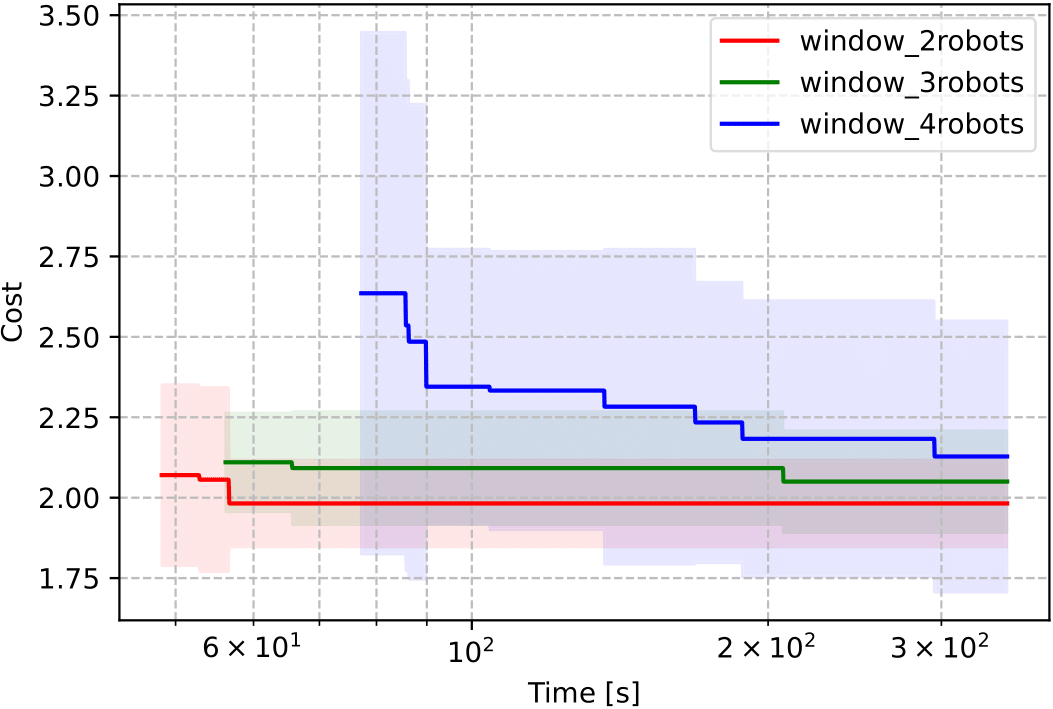}
    \caption{Anytime planning of pc-dbCBS for three different example scenarios for multirotors with payload (MP).}
    \label{fig:cost}
\end{figure}

In our work, we use the term anytime synonymously with asymptotic optimality, which is typical in the motion planning literature.The formal properties of \emph{anytime}~\cite{zilberstein1996using}, with the exception of the interruptibility before the first solution is found, are maintained. 
By decreasing $\delta$ and adding more motion primitives to the graph search step (\cref{sec:anytimeplanning}), we achieve a consistent cost reduction at each iteration, thereby demonstrating the method's asymptotic optimality as shown in \cref{fig:cost} for two, three, and four multirotors with payloads in the window environment. 
Furthermore, the optimization process operates in an anytime manner, allowing it to improve the solution over time and stops when the predefined time limit is reached. 
This ensures that the method can effectively balance solution quality and computational efficiency.

\subsection{Physical Experiments}
To validate the simulation results, we conduct real-world experiments on both platforms: multirotors with cable-suspended payload and unicycles connected by rigid rods, see \cref{fig:real_envs}. 
We describe the physical setup of each platform an the results for the problem instances.
For the multirotors with payload platform, we test scenarios with two and three multirotors. The experiments include two environments: a window-like environment requiring the robots to pass through a narrow passage and a forest-like environment with dense obstacles, similar to the scenarios that were tested for the baseline. 
For the unicycles with rigid rods platform, we conduct experiments with two and three robots in a wall-like environment. 
Here, the dynamic constraints on the angular velocities of the unicycles are $\omega^i \in [-0.5, 0.5] \ \si{rad/s}$. 
Additionally, we demonstrate a practical use-case involving three unicycles functioning as garbage collectors, as shown in the supplemental material.

\subsubsection{Multirotors with Payloads}
\renewcommand{\arraystretch}{1.0} %
\setlength{\tabcolsep}{4.5pt}       %
\begin{table}[t]
\caption{Physical experiments with multirotors with payload (MP).
Energy, tracking error, and trajectory cost over 10 trials.
}
\centering
\footnotesize
\begin{tabular}{|c||c|c||c|c||c|c|}
\hline
\multirow{2}{*}{\textbf{Environment}}
& \multicolumn{2}{c||}{\textbf{Energy} [Wh] $\downarrow$}
& \multicolumn{2}{c||}{\textbf{Error } [m] $\downarrow$}
& \multicolumn{2}{c|}{\textbf{Time} [s] $\downarrow$} \\
\cline{2-7}
& \scriptsize \textbf{Ours} & \scriptsize \textbf{BL}
& \scriptsize \textbf{Ours} & \scriptsize \textbf{BL}
& \scriptsize \textbf{Ours} & \scriptsize \textbf{BL} \\
\hline
Window, 2 robots
&
\scriptsize
{\textbf{{0.006}}\hspace{0.5em}{\tiny \textcolor{gray}{0.00}}}
&
\scriptsize
0.01 {\tiny \textcolor{gray}{0.00}}
&
\scriptsize
0.08 {\tiny \textcolor{gray}{0.05}}
&
\scriptsize
{\textbf{{0.05}}\hspace{0.5em}{\tiny \textcolor{gray}{0.03}}}
&
\scriptsize
\textbf{{4.3}}
&
\scriptsize
7.5
\\
Window, 3 robots
&
\scriptsize
{\textbf{{0.007}}\hspace{0.5em}{\tiny \textcolor{gray}{0.00}}}
&
\scriptsize
0.02 {\tiny \textcolor{gray}{0.00}}
&
\scriptsize
0.15 {\tiny \textcolor{gray}{0.06}}
&
\scriptsize
{\textbf{{0.08}}\hspace{0.5em}{\tiny \textcolor{gray}{0.04}}}
&
\scriptsize
\textbf{{4.2}}
&
\scriptsize
8.5
\\
\hline
Forest, 2 robots
&
\scriptsize
{\textbf{{0.007}}\hspace{0.5em}{\tiny \textcolor{gray}{0.00}}}
&
\scriptsize
0.01 {\tiny \textcolor{gray}{0.00}}
&
\scriptsize
0.06 {\tiny \textcolor{gray}{0.03}}
&
\scriptsize
{\textbf{{0.03}}\hspace{0.5em}{\tiny \textcolor{gray}{0.02}}}
&
\scriptsize
\textbf{{5.0}}
&
\scriptsize
8.2
\\
Forest, 3 robots
&
\scriptsize
{\textbf{{0.009}}\hspace{0.5em}{\tiny \textcolor{gray}{0.00}}}
&
\scriptsize
0.01 {\tiny \textcolor{gray}{0.00}}
&
\scriptsize
0.12 {\tiny \textcolor{gray}{0.06}}
&
\scriptsize
{\textbf{{0.07}}\hspace{0.5em}{\tiny \textcolor{gray}{0.05}}}
&
\scriptsize
\textbf{{4.5}}
&
\scriptsize
7.7
\\
\hline
\end{tabular}
\label{table3}
\end{table}
The experiments utilize Bitcraze Crazyflie 2.1 (CF) multirotors, which are small (\SI{9}{cm} rotor-to-rotor) and lightweight (\SI{34}{g}), and are commercially available. 
An existing flight controller~\cite{wahba2023efficient} is run on-board the STM32-based flight controller (168 MHz, 192 kB RAM), which also handles an extended Kalman filter for state estimation. 
For all scenarios, we use the open-sourced baseline implementation~\cite{wahba2024kinodynamic}.
On the host side, we used Crazyswarm2, an extension of Crazyswarm~\cite{preiss2017crazyswarm}, which leverages ROS~2 \cite{macenski2022robot} for commanding multiple CFs.

\subsubsection{Unicycles with Rods}
We use commercially off-the-shelf differential-drive robots of type Polulu 3pi+ 2040. The Cortex M0+ microcontroller runs MicroPython with a nonlinear controller~\cite{kanayama1990stable}.
For state estimation, we equip the robots with low-latency radios (nRF52840) and broadcast the motion capture pose at 50 Hz.
Robots are physically connected with 3D-printed rigid rods using revolute joints.

\subsubsection{Results}
We successfully execute the generated trajectories for both systems across all environments over 10 trials, except for the baseline in the three-unicycle scenario, where no successful attempts were recorded, see Tables \ref{table3}.
The trajectory times of pc-dbCBS are on average 50\% faster than the baseline and were successfully tracked by the existing controllers.
For the cable-suspended system, the executed trajectories demonstrate that our method on average consumes $50\%$ less energy than the baseline. 
As expected, the average tracking error of pc-dbCBS for the multirotors with payloads degrades at the higher execution speed due to system uncertainties, such as model mismatches and state estimation inaccuracies. However, this does not impact the success rate of the experiments.
For the unicycles, the average tracking error for pc-dbCBS three robots is $\SI{0.12}{m}$ executed in $\SI{8.9}{s}$. Similarly for two robots, the average tracking error is $\SI{0.12}{m}$ executed in $\SI{6.4}{s}$ and the baseline is $\SI{0.27}{m}$ in $\SI{13.2}{s}$. This is because the generated path from the baseline is overly curved, even though it is feasible. This suboptimality explains why the baseline results in collisions with obstacles for three robots.

\section{Conclusion}

We present pc-dbCBS, a novel kinodynamic motion planner for physically-coupled robot teams.
Algorithmically, our kinodynamic planner extends db-CBS  for physically-coupled multi-robot systems.
Our key insight is that it is beneficial to use different state representations in the same planning framework simultaneously.
Specifically, we use a stacked state representation for a discrete search over motion primitives and a minimal state representation for optimization with differential dynamic programming.

Empirically, we demonstrate that our approach computes significantly higher-quality motion plans compared to a state-of-the-art baseline on two different systems: ground robots connected via rods in a line formation, and aerial robots connected through a payload in various formations.
To this end, we derive and add a new physically-coupled example system using unicycle robots.
Our approach, pc-dbCBS, achieves a 50-60\% lower cost than the baseline, and energy consumption is reduced to 10-40\% in comparison to the baseline, depending on the number of robots and their type.

In the future, the scalability of the proposed method needs to be improved further to handle larger teams in complex environments. Moreover, a closer coupling of motion planning and controls is needed in order to plan agile trajectories that can be tracked robustly on real robots.
\printbibliography

\end{document}